\documentclass{article}
\usepackage{amsmath,graphicx,mlspconf}
\usepackage{xcolor}
\usepackage{cite}
\usepackage{amsmath,amssymb,amsfonts}
\usepackage{algorithmicx}
\usepackage{stfloats}
\usepackage{graphicx}
\usepackage{soul}
\usepackage{multirow}
\usepackage{subcaption}
\usepackage{hyperref}
\usepackage{textcomp}
\usepackage{algorithm}
\usepackage{url}
\usepackage{subcaption}
\usepackage{algpseudocode}
\usepackage{booktabs}
\usepackage{xurl}
\newcommand{\pan}[1]{{\scriptsize(#1)}}
\usepackage{xcolor}
\definecolor{LightGoldenrod}{RGB}{255, 236, 139}
\definecolor{LightYellow}{RGB}{255, 255, 150}
\definecolor{Goldenrod}{RGB}{255, 211, 0}
\definecolor{Dandelion}{RGB}{255, 185, 15}\usepackage{amsmath,graphicx,mlspconf}
\usepackage{xcolor}

\copyrightnotice{U.S.\ Government work not protected by U.S.\ copyright}

\copyrightnotice{979-8-3195-0884-3/26/\$31.00 {\copyright}2026 Crown}

\copyrightnotice{979-8-3195-0884-3/26/\$31.00 {\copyright}2026 European Union}

\copyrightnotice{979-8-3195-0884-3/26/\$31.00 {\copyright}2026 IEEE}

\toappear{2026 IEEE International Workshop on Machine Learning for Signal Processing, Sep.\ 28-- Oct.\ 1, 2026, Atlanta, USA}

\title{Jacobi-Anger Method for Deterministic Initialization in Implicit Neural Representation}

\name{Mohammed Alsakabi\textsuperscript{1}, Kejia Hu\textsuperscript{1}, John M. Dolan\textsuperscript{2}, Ozan K. Tonguz\textsuperscript{1}}
\address{\textsuperscript{1}Department of Electrical and Computer Engineering, College of Engineering\\
\textsuperscript{2}The Robotics Institute, School of Computer Science\\
Carnegie Mellon University, Pittsburgh, PA, USA\\
\texttt{\{malsakab, kejiah, jdolan, tonguz\}@andrew.cmu.edu}}

\begin{document}

\maketitle
\begin{abstract}
Existing implicit neural representation (INR) approaches suffer from stochastic initialization that does not guarantee consistent or high-quality performance across runs, with variations reaching more than 2.5 dB ($\sim$78\%) in image regression. This variation is problematic for scientific computing and simulation, where result reproducibility is crucial. To address this problem, we present Jacobi-Anger Sinusoidal Representation Network (JA-SIREN), a deterministic initialization scheme for sinusoidal networks grounded in classical spectral analysis. By computing the Discrete Sine Transform (DST) of the target signal and leveraging the Jacobi-Anger expansion, we derive deterministic weights for a two-layer sinusoidal MLP that analytically match the network's initial spectral response to the target signal. On the Kodak dataset, JA-SIREN achieves a mean PSNR of 67.18 dB, a 21.30 dB improvement over the best baseline. This is achieved with zero run-to-run variance, confirming that spectrally-informed initialization is a more effective and reproducible alternative to stochastic initialization for sinusoidal INRs. The implementation is available at the linked repository\footnote{\href{https://github.com/MohammedAlsakabi47/JA-SIREN-Initialization/tree/main}{https://github.com/MohammedAlsakabi47/JA-SIREN-Initialization/tree/main}}.
\end{abstract}

\section{Introduction}

\begin{figure}[!t]
  \centering
  \begin{subfigure}{0.8\linewidth}
    \centering
    \includegraphics[width=\linewidth]{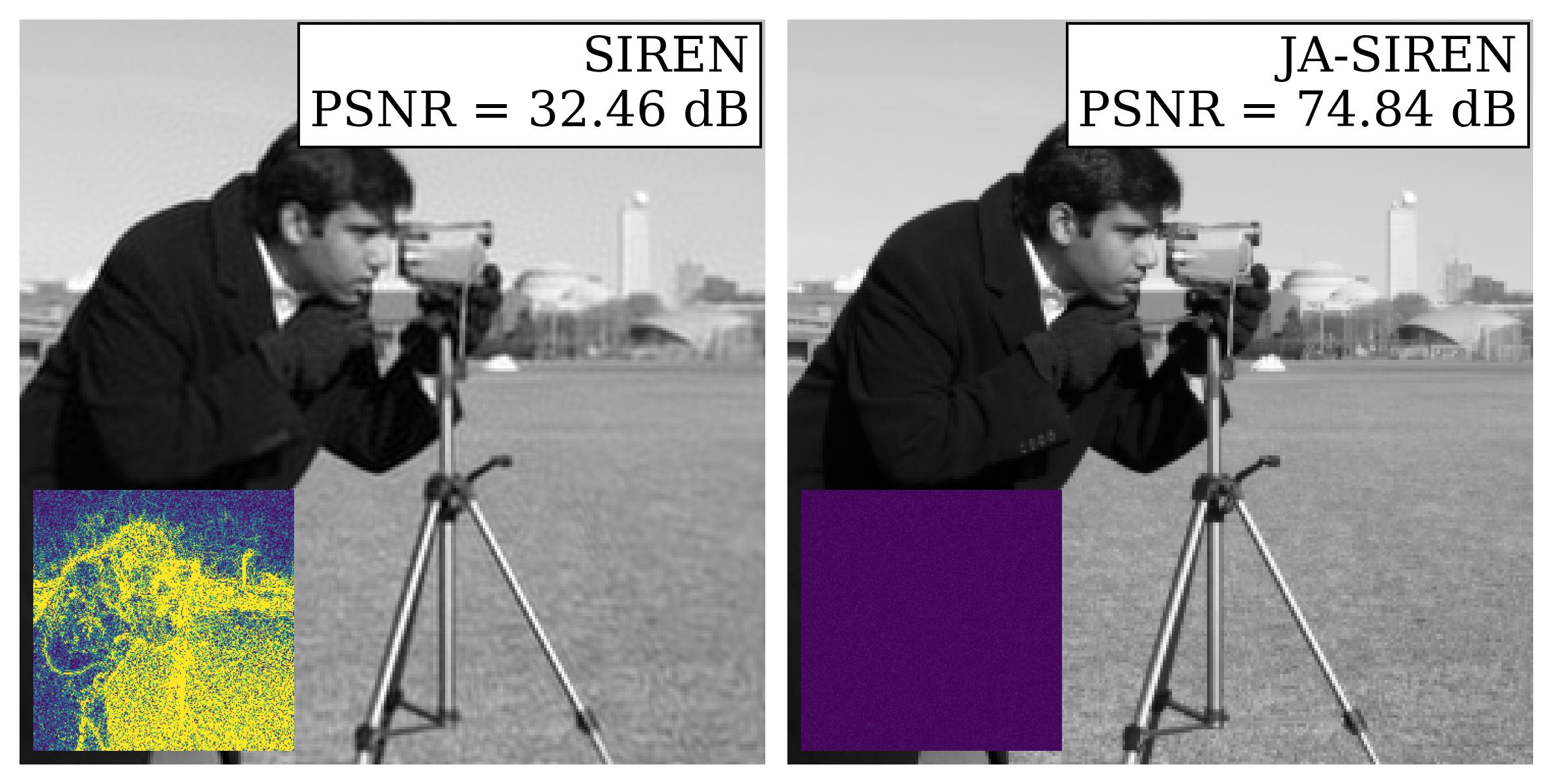}
    \caption{}
    \label{fig:comparison}
  \end{subfigure}
  
  \begin{subfigure}{\linewidth}
    \centering
    \includegraphics[width=0.8\linewidth]{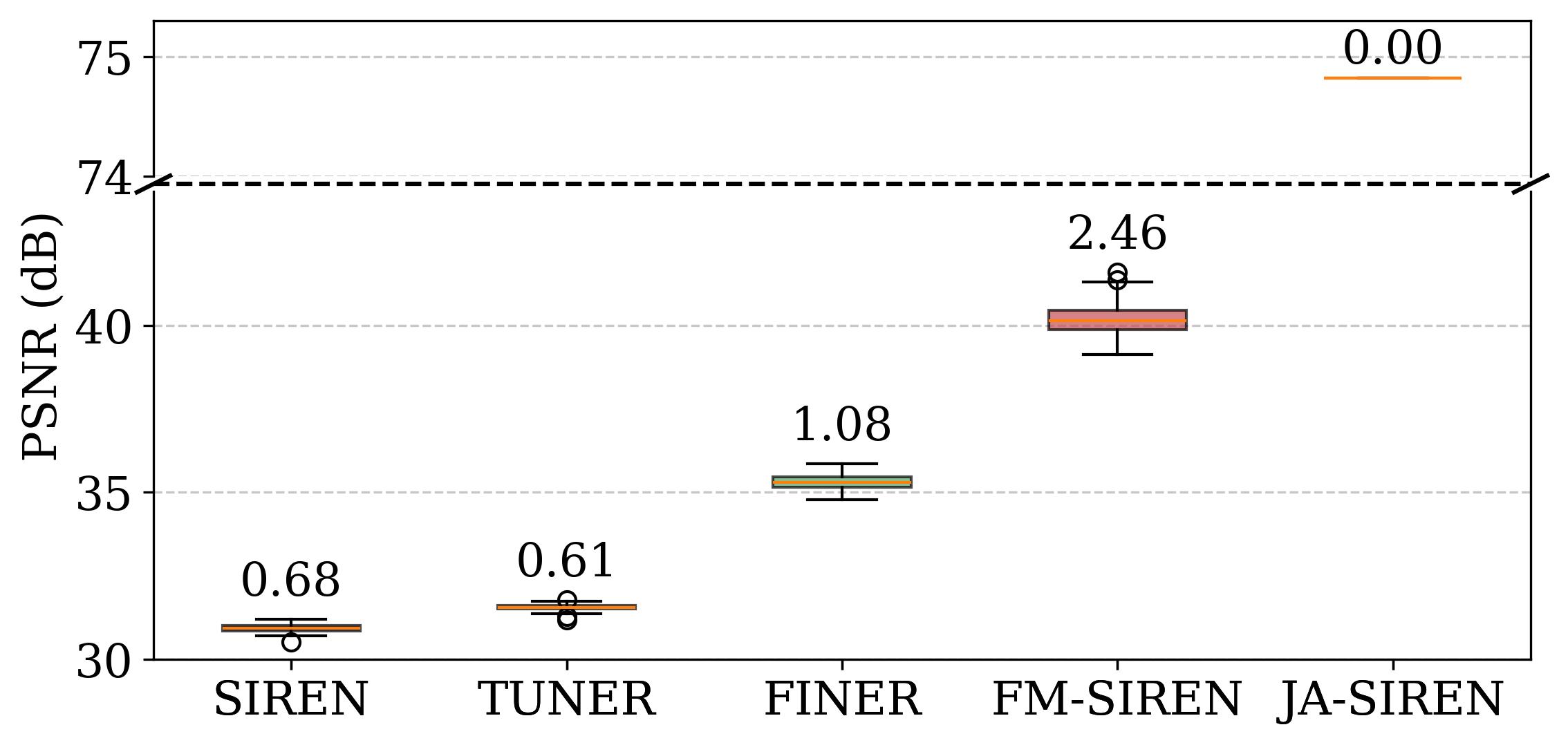}
    \caption{}
    \label{fig:reproducibility}
  \end{subfigure}
  
\caption{(a) Qualitative reconstruction results with error maps (bottom left) of SIREN and JA-SIREN for the \texttt{cameraman} image from the USC-SIPI Image Database \cite{usc-sipi-database}, where JA-SIREN achieves significantly higher fidelity. (b) PSNR results over 100 repetitions of the fitting experiment for the same image. The min-max range is shown above each box plot, confirming that JA-SIREN's deterministic initialization produces identical PSNR across all runs.}
  \label{fig:intro}
  \vspace{-20pt}
\end{figure}

Implicit neural representations (INRs) have established themselves as a compelling approach to continuous signal modeling, in which a multilayer perceptron (MLP) learns a direct mapping from input coordinates to signal values \cite{essakine2024we}. Unlike discrete representations such as grids or voxels, this coordinate-based formulation encodes a signal as the weights of a neural network, yielding representations that are inherently resolution-independent, memory-efficient, and differentiable by construction \cite{mildenhall2021nerf, martel2021acorn, xu2022signal}. This flexibility enables INRs to represent signals at arbitrary query resolutions without resampling artifacts, and to naturally incorporate gradient-based constraints such as surface normals or physical priors \cite{xu2022signal, patel2025normal}. As a result, INRs have attracted significant interest across a wide range of applications, including image and video compression, 3D shape reconstruction, novel view synthesis, and medical imaging \cite{sitzmann2019scene, mildenhall2021nerf, martel2021acorn, strumpler2022implicit}.

\par The design of INR architectures has evolved considerably, with periodic activation functions emerging as a particularly effective choice for representing signals with rich high-frequency content \cite{sitzmann2019siren, liu2024finer, saragadam2023wire, jayasundara2025mire, shah2024spder, novello2025tuning, tancik2020fourier}. Methods in this family differ in their activation design, frequency parameterization, and initialization strategies, yet they share a common underlying assumption: network weights are initialized randomly, typically sampled from a uniform, normal, or heuristically derived distribution \cite{koneputugodage2025vi}. While such schemes provide a reasonable starting point, they offer no analytical guarantee about the network's initial spectral behavior, introducing run-to-run variability in convergence speed and reconstruction quality.

\par We identify stochastic weight initialization as a fundamental limitation of existing sinusoidal INRs. Figure~\ref{fig:reproducibility} shows PSNR results of fitting the \texttt{cameraman} image over 100 repetitions \cite{usc-sipi-database}: SOTA models exhibit min-max variation of 0.61--2.46 dB across runs, confirming that initialization sensitivity is a reproducibility bottleneck. The solution lies in the Jacobi-Anger expansion, which expresses a plane wave as an infinite weighted sum of sinusoidal terms with Bessel function coefficients \cite{jacobianger}. This identity is well-suited to sinusoidal MLPs: the composition of sinusoidal activations gives rise to nested expressions of the form $\sin(c\sin(ax))$, which the expansion analytically decomposes into a weighted sum of harmonics. We exploit this connection to derive a deterministic weight initialization, analytically preconditioning the network before any gradient update with no additional parameters or hyperparameter tuning.

\par Building on this principle, we introduce \textbf{Jacobi-Anger SIREN (JA-SIREN)}, a two-layer sinusoidal MLP whose weights are initialized deterministically via the Jacobi-Anger expansion. Our contributions are: (1) We identify stochastic initialization as a fundamental reproducibility bottleneck in sinusoidal INRs and motivate a shift toward analytically grounded initialization. (2) We derive a closed-form weight initialization scheme for two-layer sinusoidal MLPs using the Jacobi-Anger expansion and Bessel functions of the first kind, requiring no random seed or hyperparameter tuning. (3) We introduce JA-SIREN and demonstrate significant reconstruction improvements over SOTA sinusoidal INRs on 1D and 2D signal fitting tasks, achieving a mean PSNR of 67.18 dB on the Kodak dataset, a 21.30 dB improvement over the best baseline.

\section{Related Work}

INR research has advanced along two axes: activation design and initialization. On the activation side, SIREN~\cite{sitzmann2019siren} established sinusoidal activations for high-frequency signal fitting; FINER~\cite{liu2024finer} extended this with variable-frequency activations; Gauss~\cite{ramasinghe2022beyond}, WIRE~\cite{saragadam2023wire}, and SPDER~\cite{shah2024spder} proposed Gaussian, Gabor, and semiperiodic alternatives respectively. Architecturally, PE~\cite{tancik2020fourier} alleviates spectral bias via sinusoidal coordinate embeddings, MIRE~\cite{jayasundara2025mire} selects per-layer activations from a dictionary, Fourier Reparameterization~\cite{shi2024improved} reformulates weights in a Fourier basis, and FM-SIREN/FM-FINER~\cite{alsakabi2025fm} assign per-neuron Nyquist-informed frequency multipliers. Despite this diversity, all rely on stochastic initialization with no analytical guarantee on the network's initial spectral state.

On initialization, Xavier~\cite{glorot2010understanding} and Kaiming~\cite{he2015delving} were derived for tanh/ReLU and transfer poorly to INRs. SAL~\cite{atzmon2020sal} and DiGS~\cite{Ben-Shabat_2022_CVPR} address geometry-driven initialization for surface reconstruction but not general INRs. SIREN's variance-preserving scheme~\cite{sitzmann2019siren} and its generalization VI$^3$NR~\cite{koneputugodage2025vi} stabilize activation statistics but remain stochastic. TUNER~\cite{novello2025tuning} and FreSh~\cite{kania2024fresh} introduce spectral initializations and scaling, yet apply uniform scaling across neurons with learnable weights. No existing method derives weights in deterministic form from the signal's analytical structure.
\begin{figure*}[t]
  \centering
  \begin{subfigure}[t]{0.29\textwidth}
    \centering
    \includegraphics[width=\linewidth]{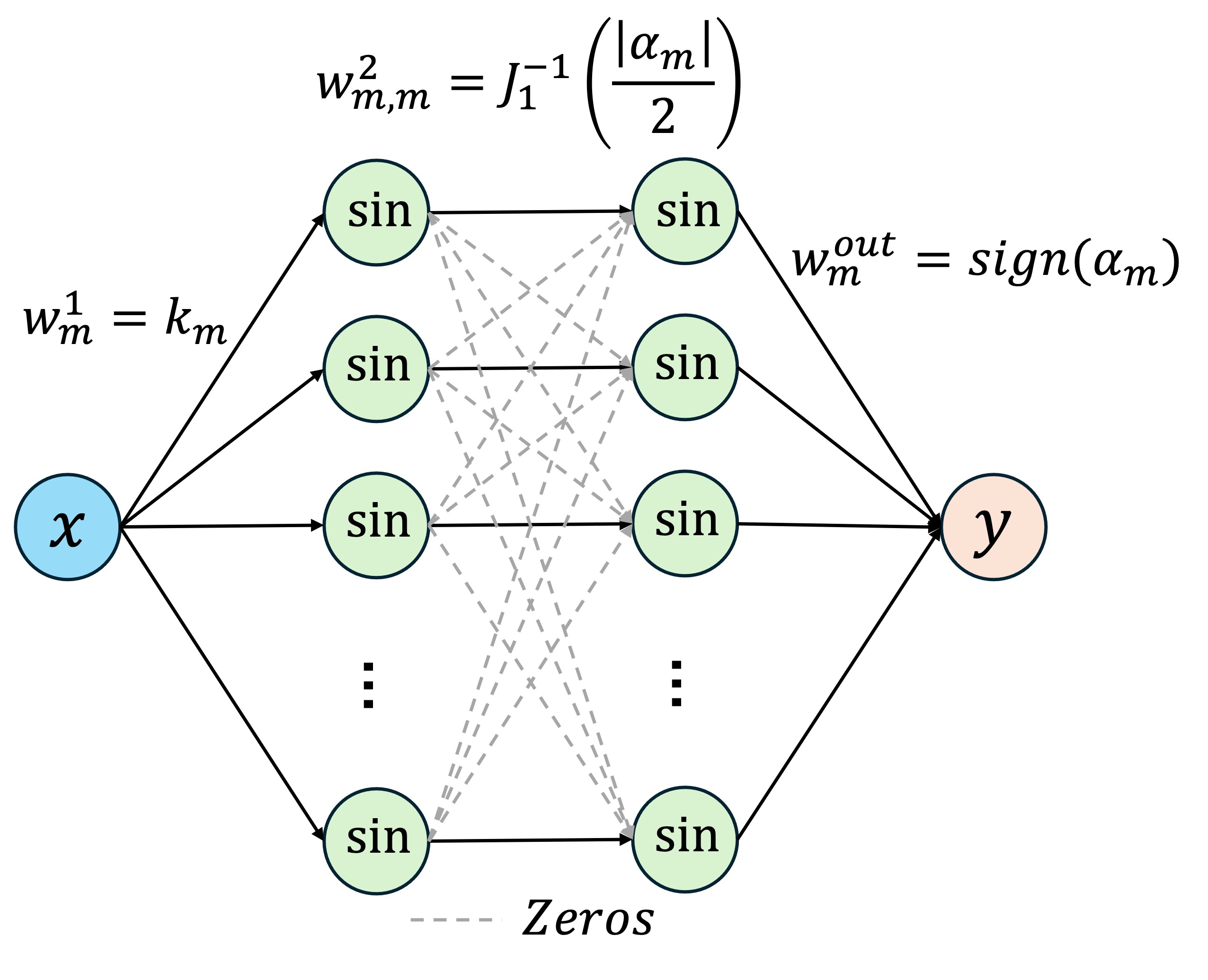}
    \caption{}
    \label{fig:sub1}
  \end{subfigure}
  \hfill
  \begin{subfigure}[t]{0.3\textwidth}
    \centering
    \includegraphics[width=\linewidth]{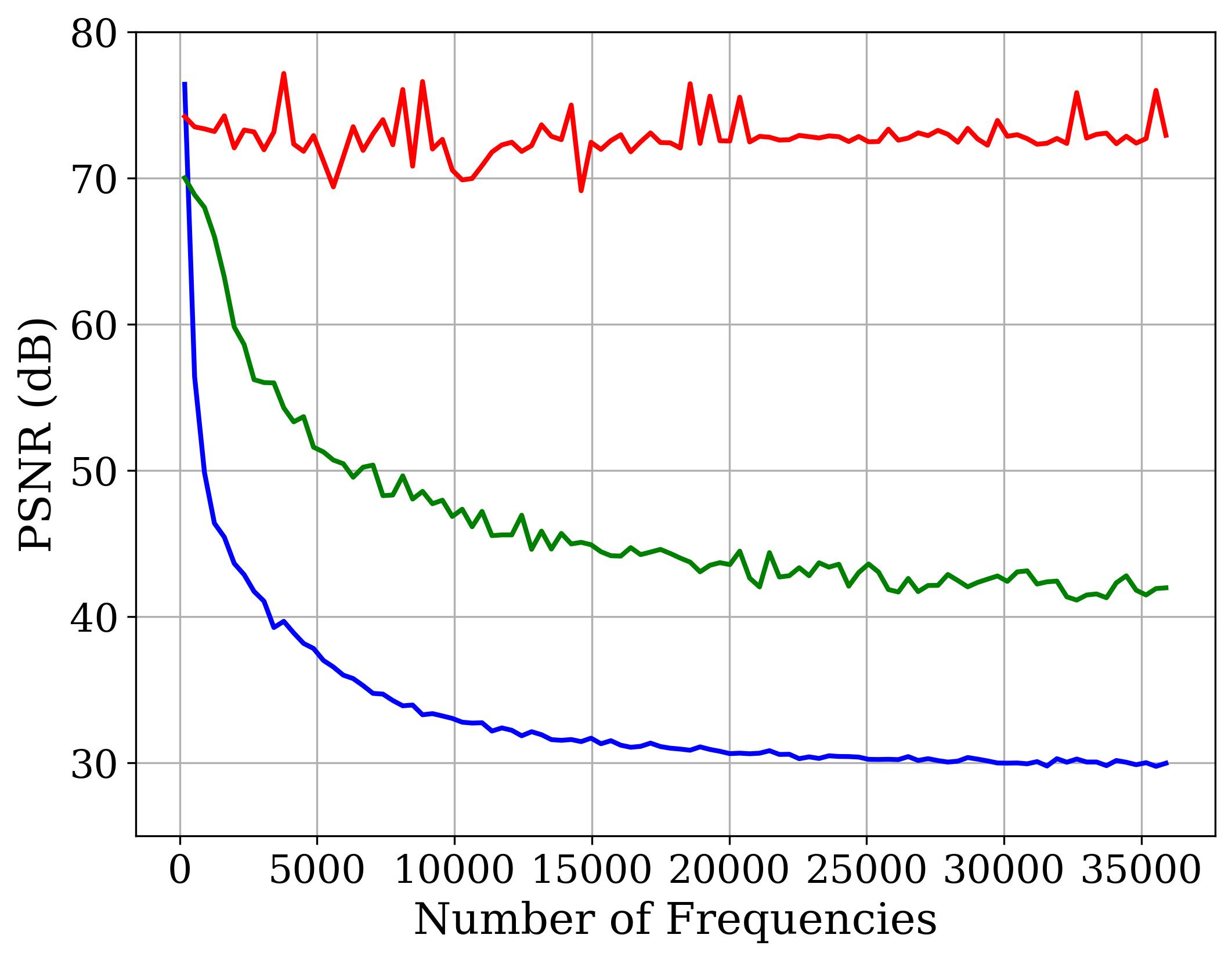}
    \caption{}
    \label{fig:sub2}
  \end{subfigure}
  \hfill
  \begin{subfigure}[t]{0.3\textwidth}
    \centering
    \includegraphics[width=\linewidth]{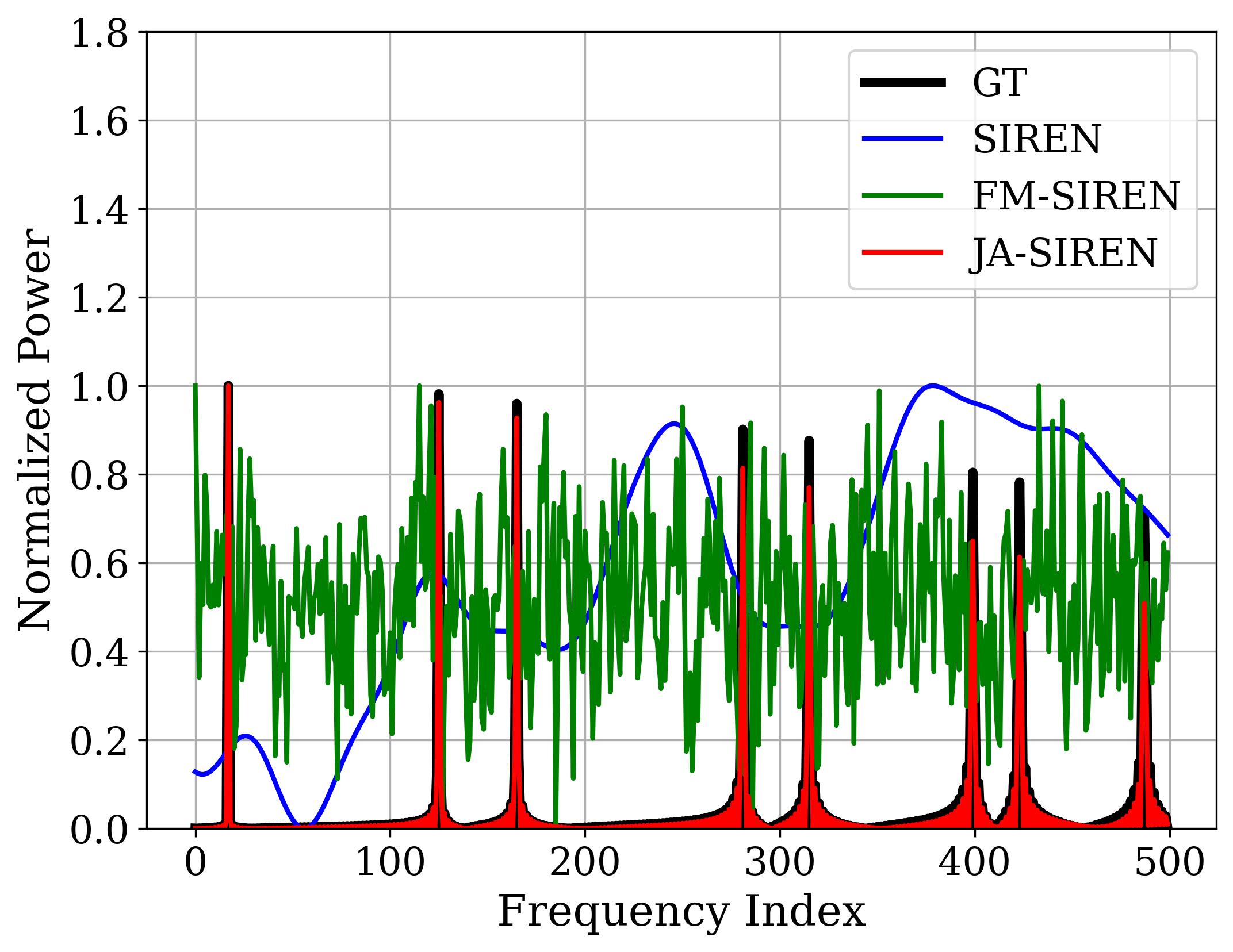}
    \caption{}
    \label{fig:sub3}
  \end{subfigure}
  \caption{(a) JA-SIREN network architecture: a two-layer sinusoidal MLP with diagonal second-layer weights, where first-layer weights $w^1_m = k_m$ encode DST frequencies, second-layer weights $w^2_m$ are set via Bessel function inversion, and output weights $w^{\text{out}}_m = \text{sign}(\alpha_m)$. (b) PSNR as a function of the number of existing DST frequency components $M$ in an input signal $f(x)$ (\texttt{camerman} in this toy example), showing convergence behavior of JA-SIREN initialization against spectrally diverse signals. (c) Normalized output power spectrum of SIREN, FM-SIREN, and JA-SIREN compared to the ground truth at initialization, demonstrating that JA-SIREN achieves the closest spectral match to the ground truth.}
  \label{fig:main}
  \vspace{-13pt}
\end{figure*}

\section{Formulation}
Our contribution builds on the Discrete Sine Transform (DST), the Jacobi-Anger expansion, and sinusoidal MLPs. In the following, we provide the necessary background for each.






\subsection{Discrete Sine Transform}
The DST represents a finite-length discrete signal as a linear combination of sinusoidal basis functions~\cite{ahmed2006discrete}. For a signal $f \in \mathbb{R}^N$, the DST-II is defined as
\begin{equation}
F_k = \sum_{n=0}^{N-1} f_n \sin\!\left(\frac{\pi(2n+1)k}{2N}\right), \quad k = 1, \ldots, N,
\label{eq:dst}
\end{equation}
where $F_k$ quantifies the contribution of the $k$-th sinusoidal basis to the signal, and $x_n = \frac{\pi(2n+1)}{2N}$ are the input coordinates. The DST-II basis functions $\phi_k(n) = \sin\!\left(\frac{\pi(2n+1)k}{2N}\right)$ form an orthogonal set, providing a complete, non-redundant spectral decomposition. Their alignment with sinusoidal activations motivates the DST as the analytical foundation for our initialization scheme.






\subsection{Jacobi-Anger Expansion}
The Jacobi-Anger expansion decomposes a complex exponential with a sinusoidal argument into an infinite series of harmonics \cite{dattoli1990theory}. Taking the imaginary part of the full expansion yields the real-valued form
\begin{equation}
    \sin(z \sin\theta) \equiv  2\sum_{n=1}^{\infty} J_{2n-1}(z)\, \sin\!\left((2n-1)\theta\right),
    \label{eq:ja_real}
\end{equation}
where only odd-order Bessel functions contribute due to the odd symmetry of sine. This identity is central to our analysis: a two-layer sinusoidal MLP with weights $a$ and $c$ computes $\sin(c\sin(ax))$, which matches the left-hand side with $z=c$ and $\theta=ax$, providing a closed-form decomposition into harmonics at odd multiples of $a$ with Bessel coefficients $J_{2n-1}(c)$.

\subsection{Sinusoidal Networks}
A sinusoidal MLP applies a sine activation to each hidden layer's linear transformation. For input $\mathbf{x} \in \mathbb{R}^{d_{in}}$, hidden layers compute $\mathbf{h}_l = \sin(\omega_0 \mathbf{W}_l \mathbf{h}_{l-1} + \mathbf{b}_l)$ with a linear output $\hat{f}(\mathbf{x}) = \mathbf{W}_L \mathbf{h}_{L-1} + \mathbf{b}_L$, where $\omega_0 > 0$ is a global frequency scale. In the special case of $L=2$ hidden layers and scalar input $x \in \mathbb{R}$, each neuron $m$ computes $\sin(c\sin(ax))$ with $a = \omega_0 w_{1,m}$ and $c = \omega_0 w_{2,m}$, precisely the structure analyzed by the Jacobi-Anger expansion.

\begin{algorithm}[t]
\caption{JA-SIREN Initialization}
\label{alg:ja_siren}
\begin{algorithmic}[1]
\Require Raw signal $s$, network width $M$
\Ensure $\{w_m^1\}_{m=1}^{M}$, $\mathbf{W}^2 \in \mathbb{R}^{M \times M}$ (diagonal), $\{w_m^{out}\}_{m=1}^{M}$
\State Flatten $s \to v \in \mathbb{R}^N$; compute grid $x_n = \frac{\pi(2n+1)}{2N}$
\State Compute DST-II coefficients $V$; let $\{k_m\}$ be top-$M$ magnitude indices
\State Set $\alpha_m \leftarrow V_{k_m}/N$, $\;w_m^1 \leftarrow k_m$, $\;w_m^{out} \leftarrow \operatorname{sign}(\alpha_m)$
\State $\mathbf{W}^2_{m,m} \leftarrow J_1^{-1}(|\alpha_m|/2)$ if $|\alpha_m|/2 \leq J_1^{\max}$, else $|\alpha_m|/2$ 
\State Set all biases to zero
\State \Return $\{w_m^1\}$, $\mathbf{W}^2$, $\{w_m^{out}\}$
\end{algorithmic}
\end{algorithm}

\section{Proposed Method: JA-SIREN}
Consider a sinusoidal MLP with two hidden layers, width $M$, and a linear output layer. We flatten the input data into a 1D signal and compute the DST over the discrete coordinate grid $\{x_n\}_{n=0}^{N-1}$, where $x_n = \frac{\pi(2n+1)}{2N}$. Retaining the top $M$ frequency components by magnitude yields:
\begin{equation}
    \hat{f}(x_n) = \sum_{m=1}^{M} \alpha_m \sin(k_m x_n)
    \label{eq:approx}
\end{equation}
where $\alpha_m$ is the signed amplitude and $k_m$ the DST frequency index of the $m$-th dominant component.

We impose a diagonal second-layer weight matrix so that each neuron $m$ receives input only from its corresponding first-layer neuron, giving:
\begin{equation}
    y_n = \sum_{m=1}^{M} w_m^{out} \sin(w_{m}^{2} \sin(w_m^1 x_n))
    \label{eq:mlp_output}
\end{equation}

Equating each neuron's output to its corresponding DST term in \eqref{eq:approx} and substituting $w_m^1 = k_m$ yields:

\begin{equation}
    w_m^{out} \sin(w_{m}^{2} \sin(k_m x_n)) \equiv \alpha_m \sin(k_m x_n)
    \label{eq:equivalence}
\end{equation}

Applying the Jacobi-Anger identity \eqref{eq:ja_real} and retaining only the first-order term ($n=1$) (justified by the rapid decay of $J_{2n-1}(z)$ for $n>1$) gives the matching condition:
\begin{equation}
    2 w_m^{out} J_1(w_{m}^{2}) = \alpha_m
    \label{eq:matching}
\end{equation}

Since $J_1(z) > 0$ for $z \in (0, j_{1,1})$ where $j_{1,1} \approx 3.83$, we absorb the sign of $\alpha_m$ into the output weight by setting $w_m^{out} = \text{sign}(\alpha_m)$, and solve for $w_m^2$:

\begin{equation}
    w_{m}^{2} = J_1^{-1}\!\left(\frac{|\alpha_m|}{2}\right)
    \label{eq:w2_solution}
\end{equation}

where $J_1^{-1}$ is computed via root-finding on $(0, j_{1,1})$. Together with $w_m^1 = k_m$ and $w_m^{out} = \text{sign}(\alpha_m)$, this yields a fully deterministic, closed-form initialization requiring no random seed or hyperparameter tuning. Algorithm~\ref{alg:ja_siren} summarizes the full procedure.

As shown in Figure~\ref{fig:main}a, JA-SIREN is a two-layer sinusoidal MLP with diagonal second-layer weights derived from the DST and Jacobi-Anger expansion. Figure~\ref{fig:main}b shows reconstruction quality as the number of DST frequency components increases, and Figure~\ref{fig:main}c validates the spectral matching property at initialization, where JA-SIREN's output power spectrum most closely follows the ground truth.
\section{Experiments}
\vspace{-20pt}
\begin{figure*}[t]
    \centering
    \begin{subfigure}[b]{0.17\linewidth}
        \includegraphics[width=\textwidth]{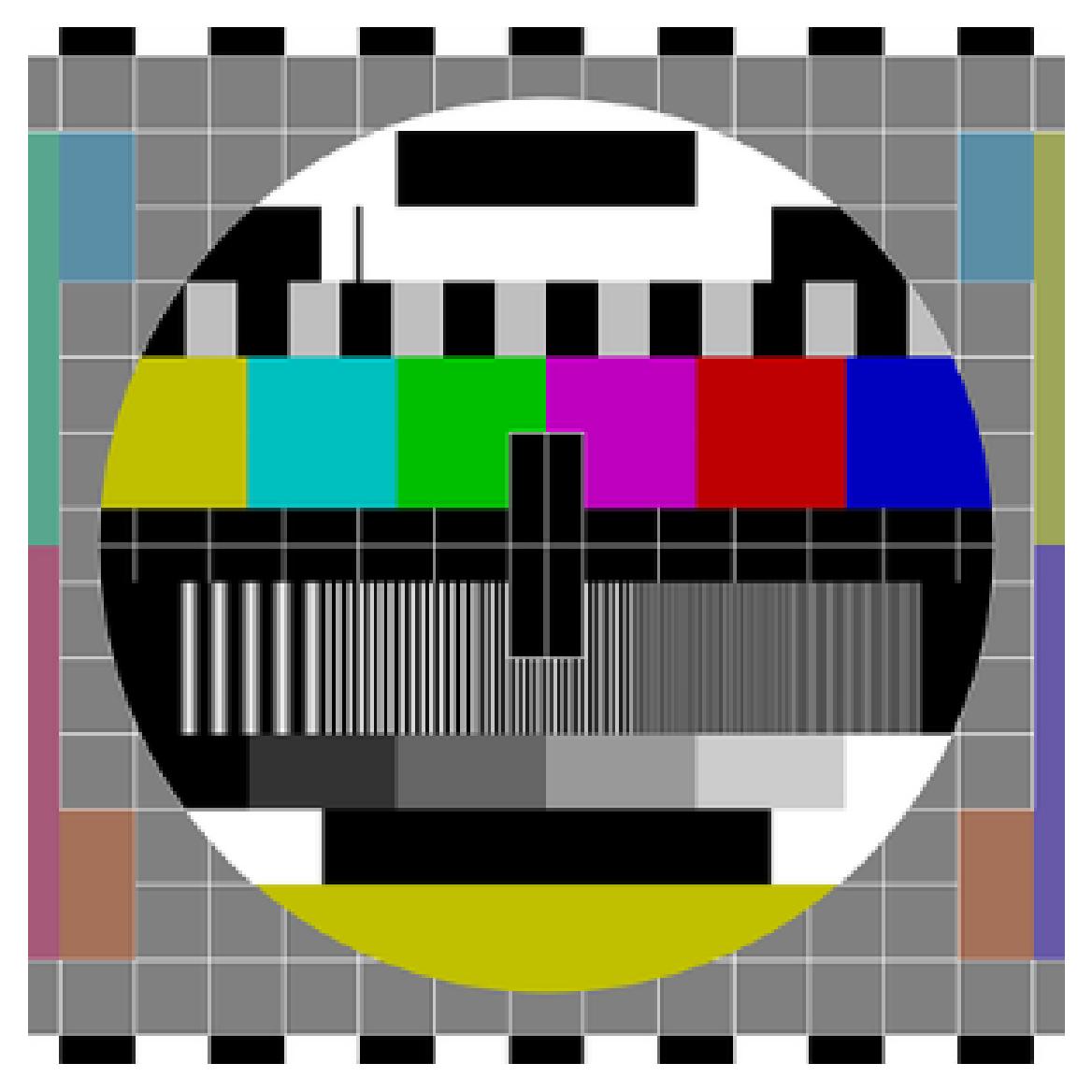}
        \caption{Ground Truth}
        \label{fig:sub1}
    \end{subfigure}
    \begin{subfigure}[b]{0.17\linewidth}
        \includegraphics[width=\textwidth]{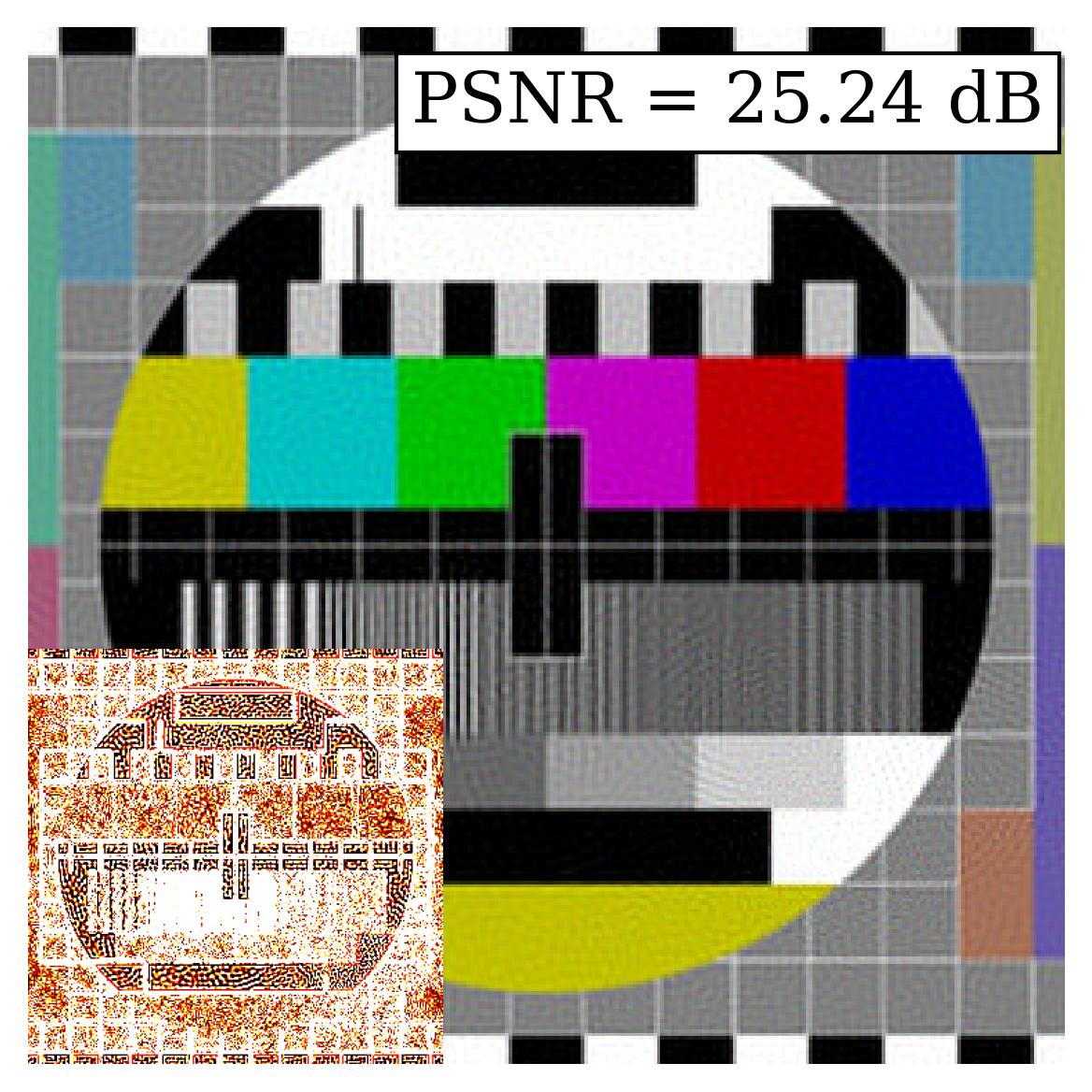}
        \caption{SIREN}
        \label{fig:sub1}
    \end{subfigure}
    \begin{subfigure}[b]{0.17\linewidth}
        \includegraphics[width=\textwidth]{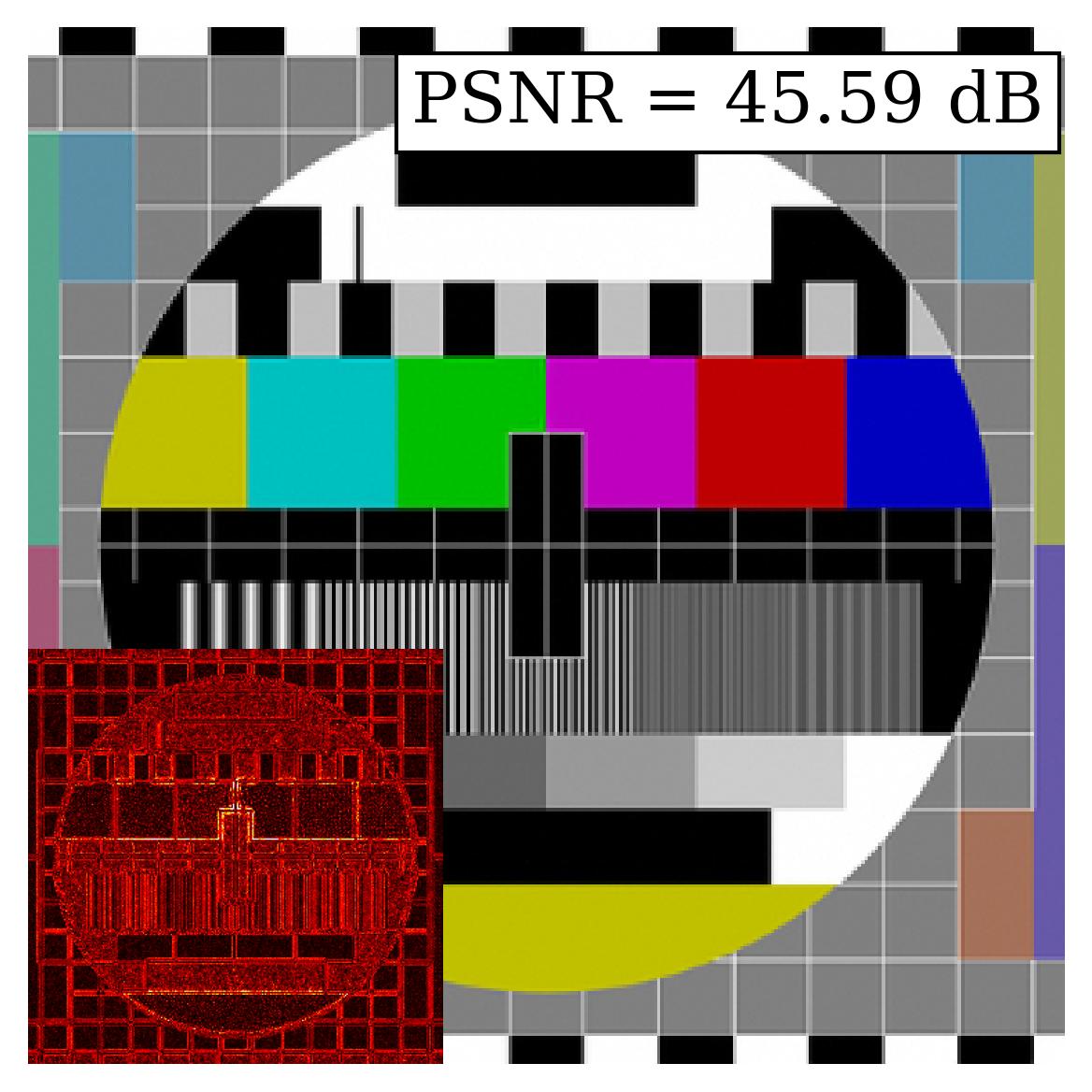}
        \caption{FM-SIREN}
        \label{fig:sub2}
    \end{subfigure}
    \begin{subfigure}[b]{0.17\linewidth}
        \includegraphics[width=\textwidth]{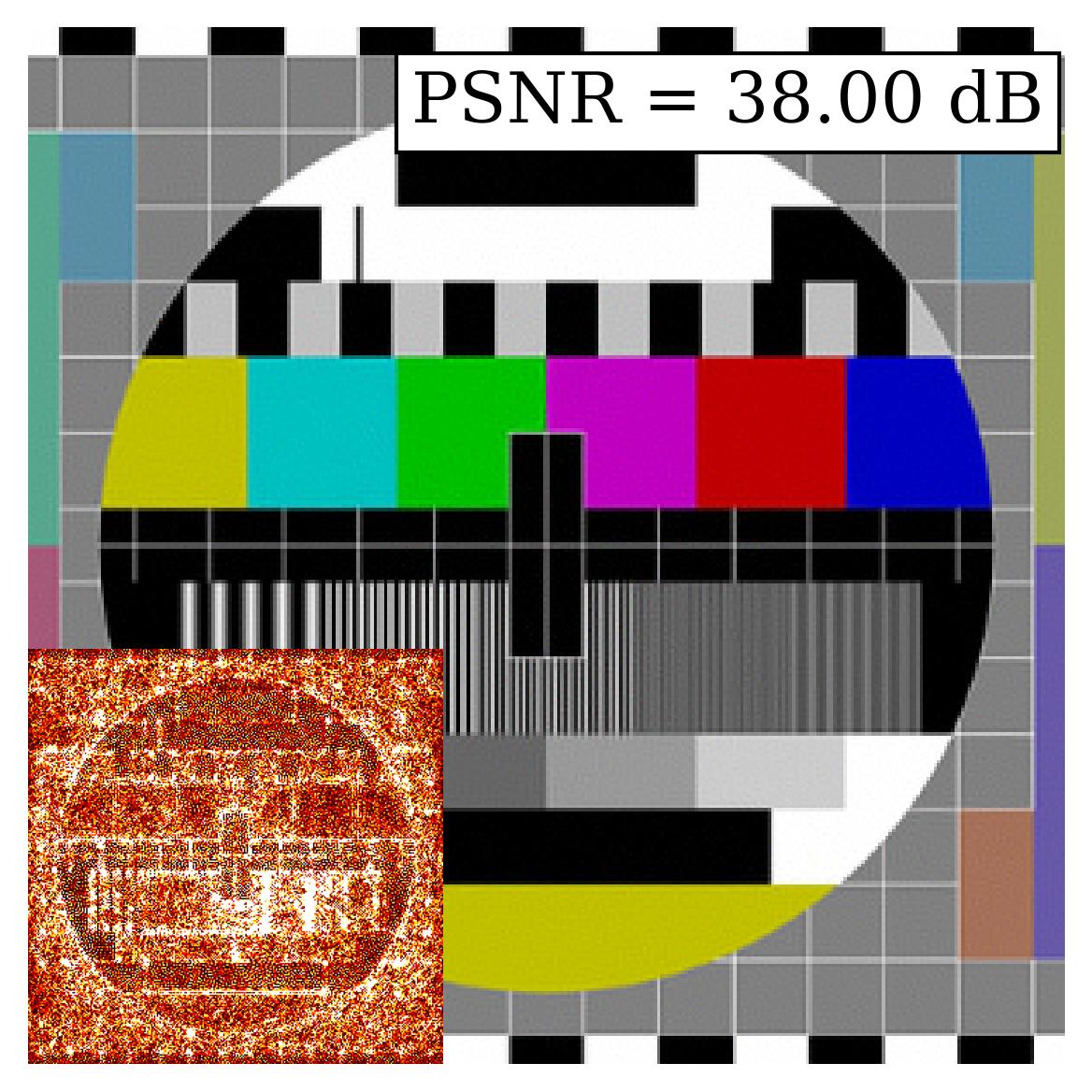}
        \caption{FINER}
        \label{fig:sub3}
    \end{subfigure}
    \begin{subfigure}[b]{0.17\linewidth}
        \includegraphics[width=\textwidth]{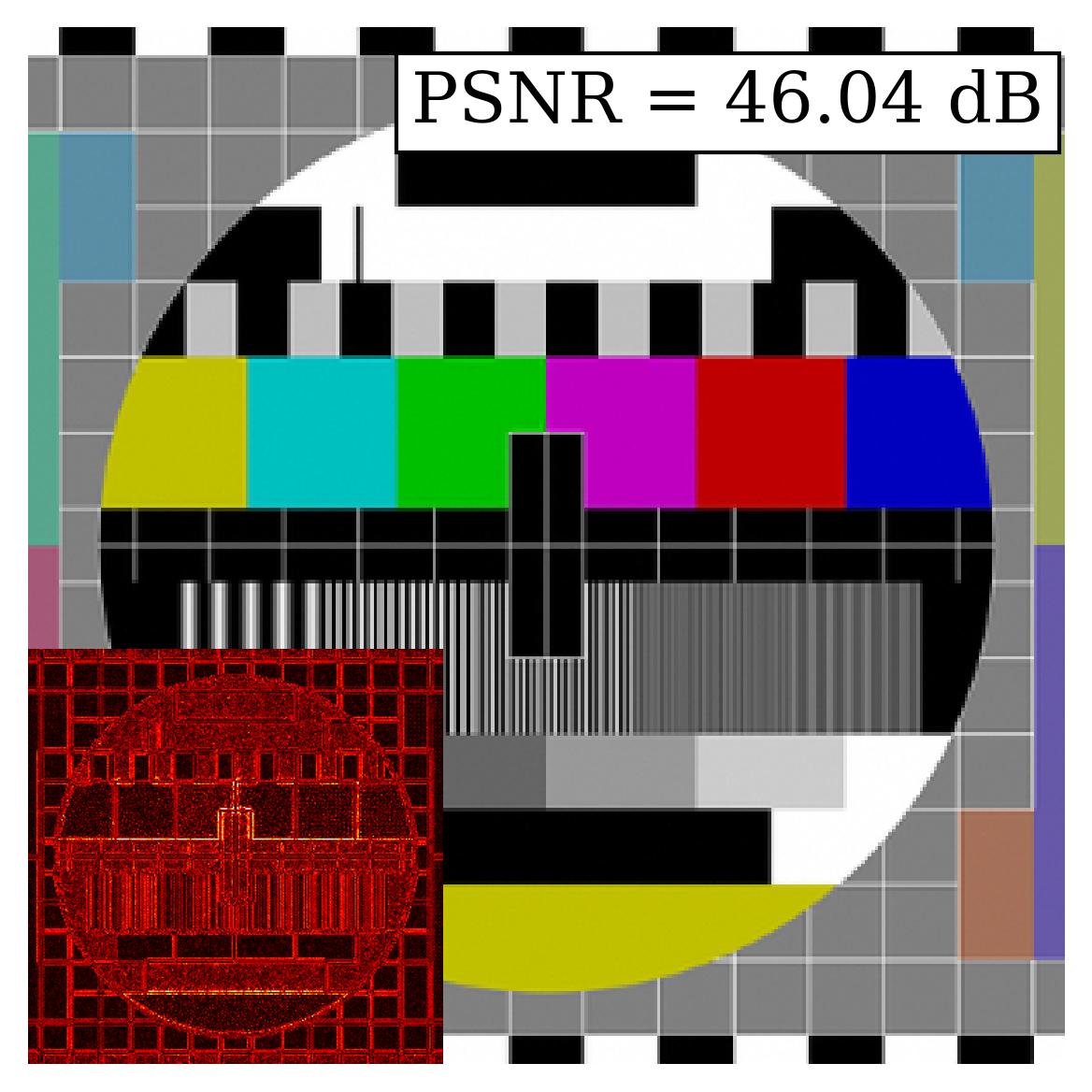}
        \caption{FM-FINER}
        \label{fig:sub4}
    \end{subfigure}

    \vspace{0.5em}

    \begin{subfigure}[b]{0.17\linewidth}
        \includegraphics[width=\textwidth]{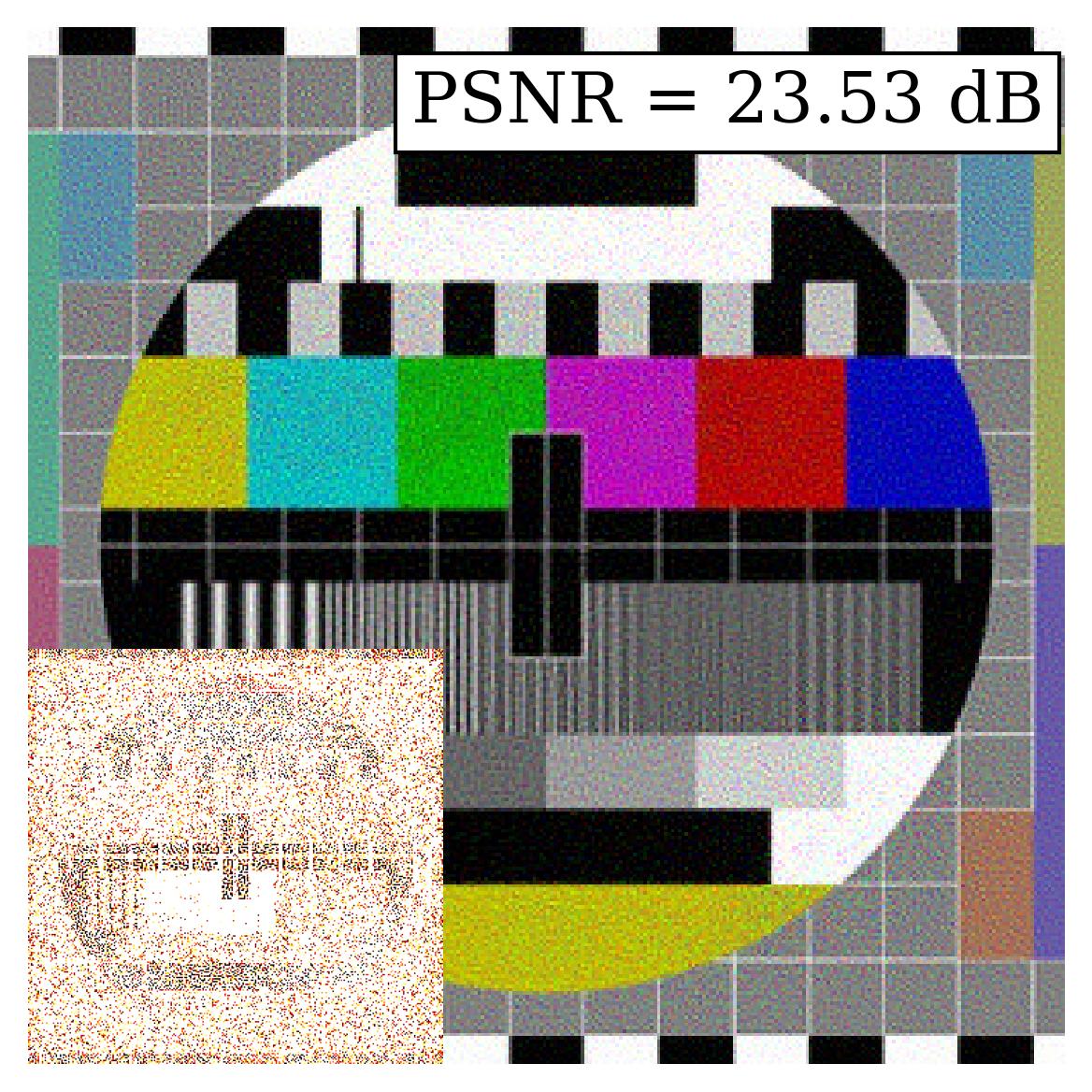}
        \caption{SPDER}
        \label{fig:sub6}
    \end{subfigure}
    \begin{subfigure}[b]{0.17\linewidth}
        \includegraphics[width=\textwidth]{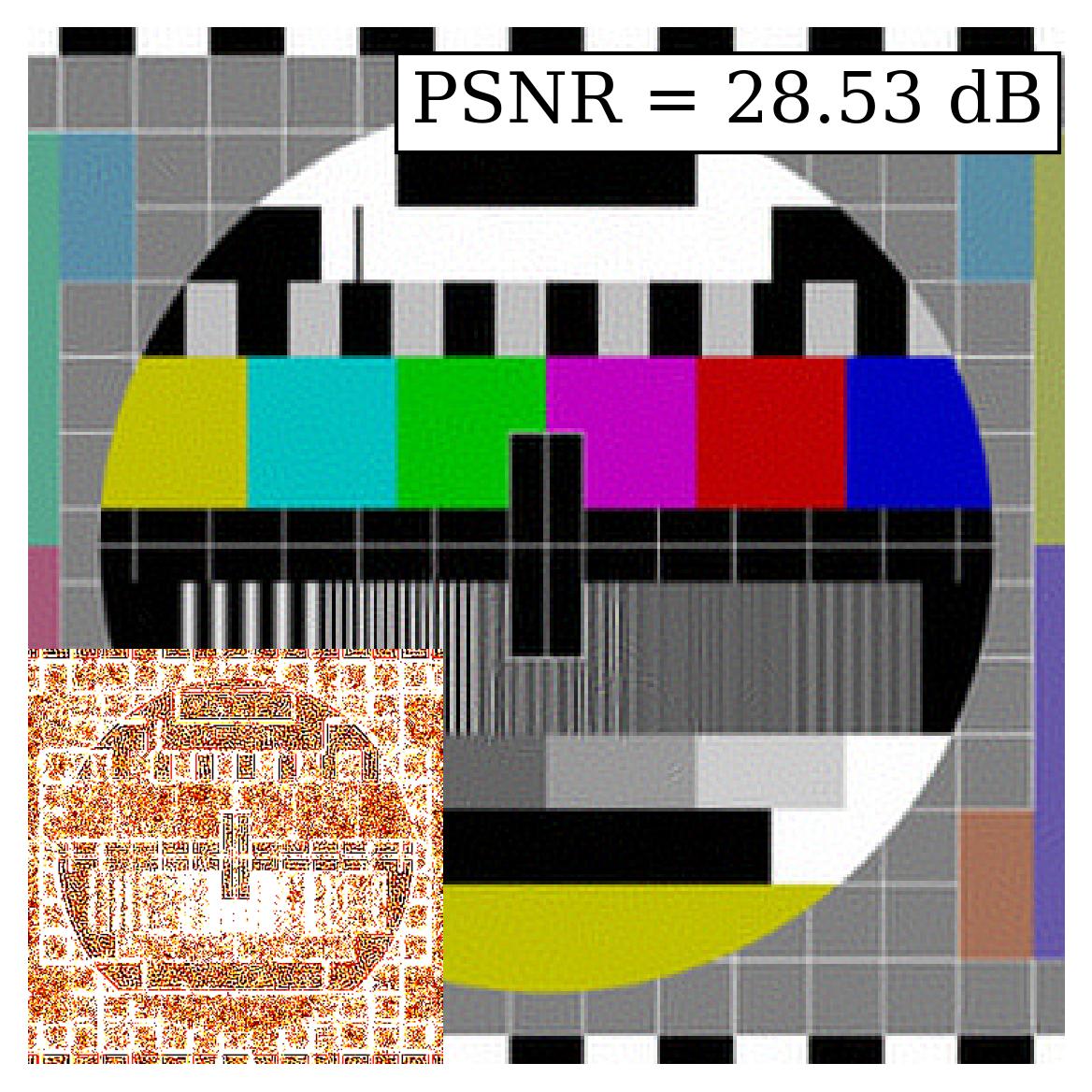}
        \caption{WIRE}
        \label{fig:sub7}
    \end{subfigure}
    \begin{subfigure}[b]{0.17\linewidth}
        \includegraphics[width=\textwidth]{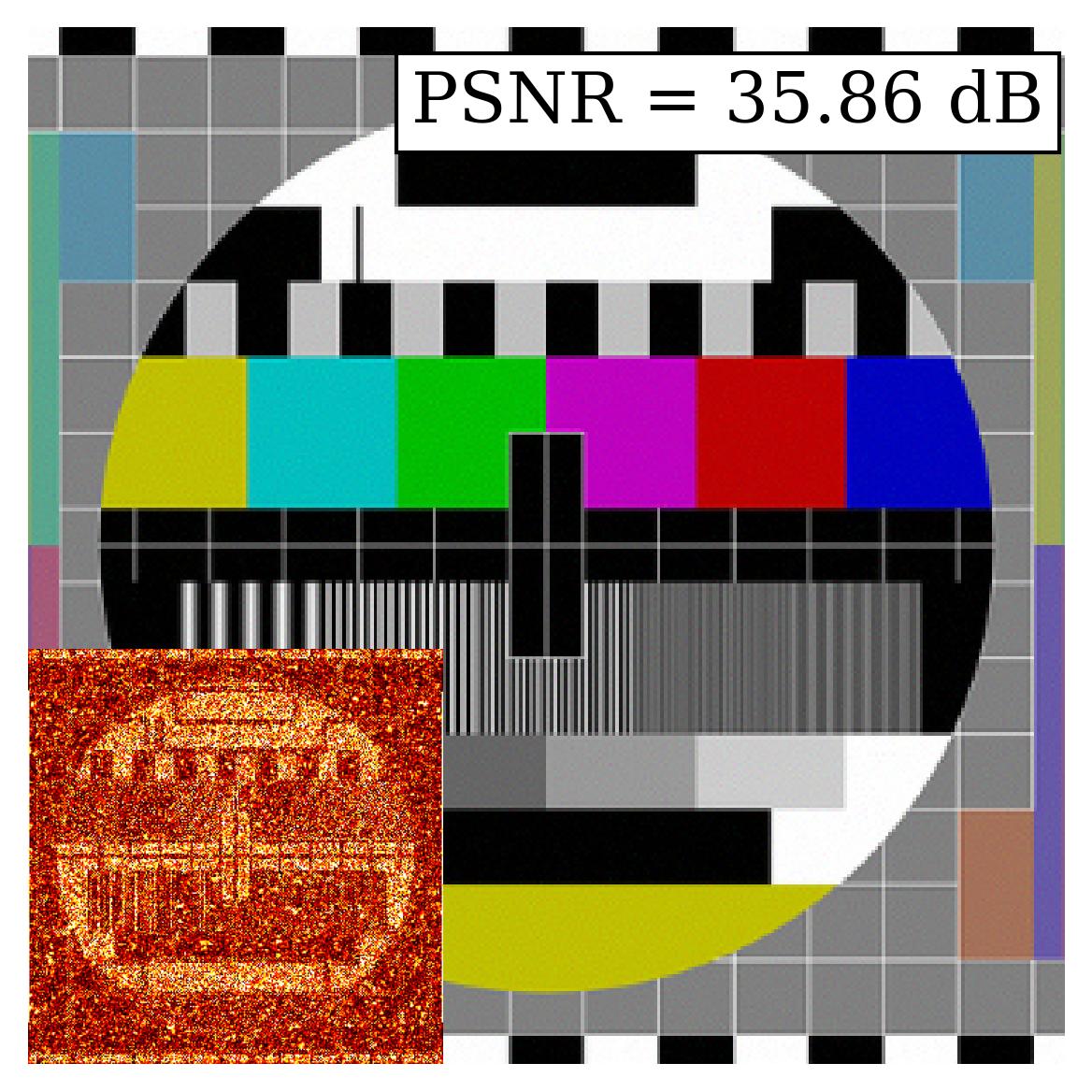}
        \caption{TUNER}
        \label{fig:sub8}
    \end{subfigure}
    \begin{subfigure}[b]{0.17\linewidth}
        \includegraphics[width=\textwidth]{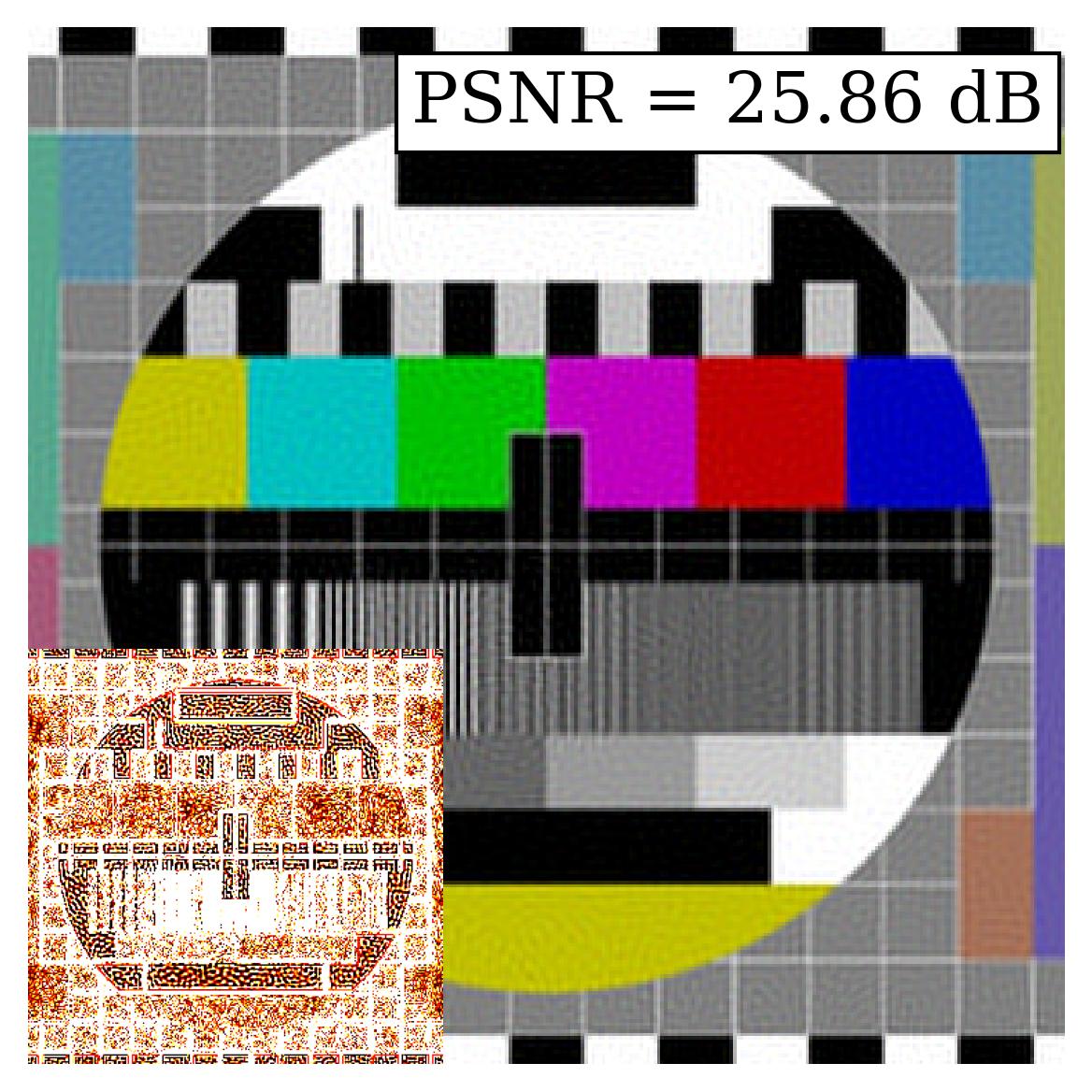}
        \caption{FreSh}
        \label{fig:sub9}
    \end{subfigure}
    \begin{subfigure}[b]{0.17\linewidth}
        \includegraphics[width=\textwidth]{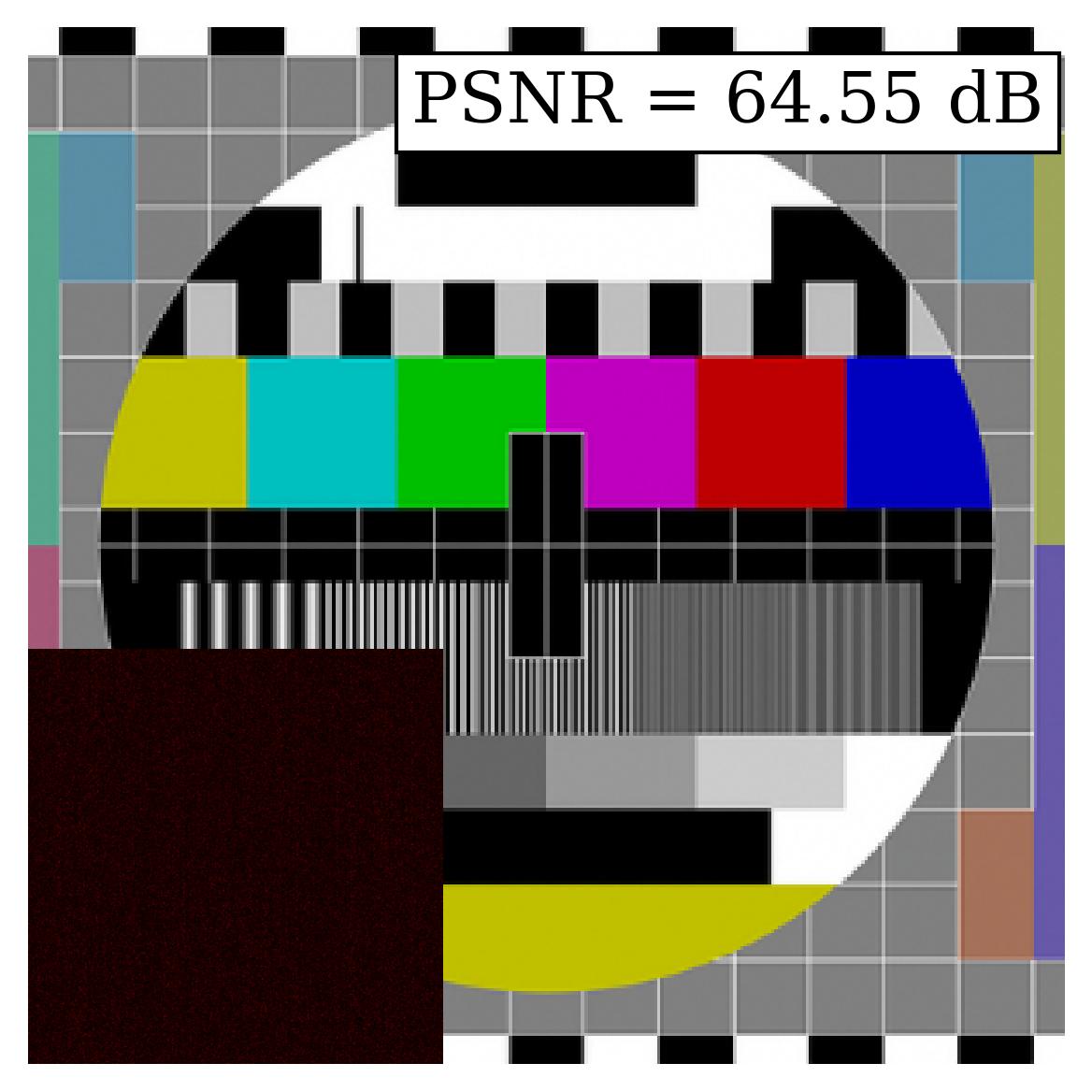}
        \caption{JA-SIREN}
        \label{fig:sub10}
    \end{subfigure}

    \caption{Qualitative image reconstruction results of the \texttt{PCP} image \cite{philips_circle_wiki} using two-layer networks. The PSNR (dB) of each reconstruction is reported in the top-right corner and the error map in the bottom-left corner of its subfigure. JA-SIREN achieves the highest PSNR values and produces visibly sharper reconstructions compared to the baselines. While FM-FINER attains the best performance among the baseline methods, its reconstructions appear noisier than JA-SIREN, as highlighted in the error map. Observe that JA-SIREN improves the PSNR of FM-FINER, which is the closest competitor, by about 19 dB.}
    \label{fig:experimantal_result}
    \vspace{-10pt}
\end{figure*}

\begin{table}[t]
    \centering
    \caption{Reconstruction performance on the Kodak dataset~\cite{kodak} and audio regression on the Spoken Wikipedia dataset~\cite{spokenwikipedia}. Note that TUNER is not designed to fit 1D audio signals. Standard deviations are shown in parentheses. \colorbox{Dandelion}{Best}, \colorbox{Goldenrod}{Second Best} and \colorbox{LightYellow}{Third Best} are highlighted.}
    \label{tab:resutls}
    \resizebox{\columnwidth}{!}{%
    \begin{tabular}{l*{2}{c}cc}
        \toprule
        \multirow{2}{*}{\textbf{Model}} & \multicolumn{2}{c}{\textbf{Image}} & \textbf{Audio} & \multirow{2}{*}{\textbf{Fitting Time (min)}} \\
        \cmidrule(lr){2-3} \cmidrule(lr){4-4}
        & \textbf{PSNR (dB) $\uparrow$} & \textbf{SSIM $\uparrow$} & \textbf{MSE $\times 10^{-3}\downarrow$} & \\
        \midrule
        SIREN \cite{sitzmann2019scene}    & $28.00$ \pan{4.32} & $0.7882$ \pan{0.16571} & $0.513$ \pan{0.012} & $3.63$ \\
        FINER \cite{liu2024finer}         & $37.16$ {\pan{1.50}} & $0.9614$ \pan{0.01121} & $0.263$ \pan{0.022} & $3.12$ \\
        WIRE \cite{saragadam2023wire}     & $31.88$ \pan{2.29} & $0.9082$ \pan{0.02142} & $5.812$ \pan{0.001} & $4.10$ \\
        Gauss \cite{ramasinghe2022beyond} & $25.06$ \pan{2.52} & $0.6657$ \pan{0.10078} & $5.948$ \pan{0.000} & $11.15$ \\
        PE \cite{tancik2020fourier}       & $35.82$ \pan{2.56} & $0.9583$ \pan{0.00968} & $6.884$ \pan{0.000} & $3.31$ \\
        TUNER \cite{novello2025tuning}    & $36.74$ \pan{2.75} & $0.9806$ \pan{0.01002} & - & - \\
        SPDER \cite{shah2024spder}        & $22.28$ \pan{4.46} & $0.5676$ \pan{0.16491} & $1.476$ \pan{2.676} & $8.33$ \\
        FM-SIREN  \cite{alsakabi2025fm}   & \colorbox{LightYellow}{44.33 \pan{2.33}} & \colorbox{LightYellow}{0.9918 \pan{0.00366}} & \colorbox{LightYellow}{$0.047$ \pan{0.135}} & $3.12$ \\
        FM-FINER \cite{alsakabi2025fm}    & \colorbox{Goldenrod}{45.88 \pan{3.37}} & \colorbox{Goldenrod}{0.9940 \pan{0.00326}} & \colorbox{Goldenrod}{$0.040$ \pan{0.258}} & $3.63$ \\
        \midrule
        \textbf{JA-SIREN}                 & \colorbox{Dandelion}{67.18 \pan{0.92}} & \colorbox{Dandelion}{0.9999 \pan{0.00003}} & \colorbox{Dandelion}{$0.030$ \pan{0.040}} & $3.17$ \\
        \midrule
        \textbf{Improvement}              & $+21.30$ \pan{$-$2.45} & $+0.0059$ \pan{$-$0.00323} & $-0.010$ \pan{-0.218} & -- \\
        \bottomrule
    \end{tabular}}
    \vspace{-10pt}
\end{table}

\subsection{Experimental Setup}
We used PyTorch \cite{paszke2019pytorch} with the Adam optimizer \cite{bottou2018optimization} and a StepLR scheduler that decayed the learning rate by a factor of 0.5 every 200 epochs, and we trained all models for 1000 epochs. All experiments were conducted on an NVIDIA H200 (80GB VRAM, 256GB RAM). To ensure reproducibility across different types of GPUs, we disabled \texttt{TF32} for both matrix multiplications and \texttt{cuDNN} operations, and enforced deterministic \texttt{cuDNN} behavior with benchmarking disabled.

\subsection{Experimental Results}
We evaluated a 512-neuron two-layer sinusoidal MLP initialized with JA-SIREN on all images in the Kodak Lossless True Color Image Suite \cite{kodak}, as well as the Philips Circle Pattern (\texttt{PCP}) \cite{philips_circle_wiki}, which contains both low- and high-frequency features (i.e., smooth regions and sharp edges), making it well-suited for testing reconstruction fidelity across the frequency spectrum. The performance is assessed using two standard metrics: Peak Signal-to-Noise Ratio (PSNR) and the Structural Similarity Index Measure (SSIM) \cite{hore2010image}. As shown in Figure \ref{fig:experimantal_result}, JA-SIREN achieves remarkable reconstruction fidelity on the PCP image, obtaining 64.55 dB compared to 46.04 dB for FM-FINER, corresponding to 71$\times$ improvement as a linear factor. On the Kodak dataset, JA-SIREN achieves a mean PSNR of 67.18 dB, representing a 21.30 dB improvement over the best baseline, FM-FINER (45.88 dB), which corresponds to a 134.9$\times$ improvement. Furthermore, JA-SIREN achieves a mean SSIM of 0.9999, which is arguably lossless reconstruction. As shown in Table \ref{tab:resutls}, JA-SIREN also exhibits the lowest standard deviations in both PSNR and SSIM across the Kodak dataset, indicating consistent performance across diverse inputs.

We further evaluated 256-neuron JA-SIREN on 1D audio regression using the Spoken English Wikipedia dataset \cite{spokenwikipedia}, a diverse collection of audio recordings spanning a wide range of topics. Each audio clip was resampled to 4 kHz and the first 5 seconds were retained, yielding signals of length 20,000. As shown in Table~\ref{tab:resutls}, JA-SIREN achieves a mean MSE of $0.030 \times 10^{-3}$, outperforming the best baseline, FM-FINER ($0.040 \times 10^{-3}$), demonstrating that the deterministic spectral initialization generalizes beyond image regression to 1D signals, which exhibit a significantly wider frequency spectrum than natural images.

\vspace{-9pt}

\section{Discussion}
The experimental results provide strong evidence that deterministic, spectrally-informed initialization is a more effective starting point than stochastic initialization for sinusoidal INRs. By grounding the initialization in the DST coefficients of the target signal, JA-SIREN replaces an arbitrary spectral configuration with one that is analytically matched to the input signal, yielding PSNR gains exceeding 21 dB over the best stochastic baseline on the Kodak dataset. Although a favorable random seed could in principle yield comparable performance, the probability of drawing one is essentially zero, since the optimal first-layer weights $w^1_m = k_m$ are specific integer-valued DST frequency indices while stochastic baselines sample from a continuous uniform distribution. This is empirically confirmed by the 100-repetition experiment in Figure~\ref{fig:intro}b, where no stochastic baseline approached the performance of JA-SIREN's deterministic initialization. The consistently low standard deviation in both PSNR and SSIM confirms that this per-signal adaptation scales with the signal's structure, maintaining high reconstruction fidelity across diverse inputs, from smooth low-frequency images to sharp high-frequency edges, and further generalizes to 1D signals such as audio, which exhibit a significantly wider frequency spectrum than natural images.

Beyond reconstruction quality, deterministic initialization has significant implications for scientific reproducibility. Unlike stochastic methods, whose results vary across runs and hardware, JA-SIREN produces identical outputs given the same input, regardless of random seed. This property is particularly valuable in fields where reproducibility is critical, such as data compression, medical imaging, and physical simulation, and further extends to scientific computing applications where INRs are used to approximate solutions to partial differential equations. 

While reconstruction quality beyond 40~dB may already appear visually excellent, the 60+~dB regime is empirical evidence that deterministic spectral matching works as a principled initialization strategy; the magnitude of the gap reflects how far stochastic methods are from the analytically optimal configuration, and applications such as lossless compression, medical imaging, and PDE-based scientific computing demand numerical precision beyond perceptual thresholds, where JA-SIREN's deterministic guarantees are fundamentally out of reach for stochastic methods.

\section{Conclusion and Future Work}

In this paper, we introduced JA-SIREN, a new initialization scheme for 
sinusoidal implicit neural representations. Our proposed scheme is grounded in the spectral response of the DST and the expansion of its dominant frequency components via the Jacobi-Anger expansion. This initialization is deterministic and adaptive to the input signal, making it well-suited for rigorous analysis and result reproducibility --- properties required in scientific computing applications. Moreover, JA-SIREN demonstrates significant reconstruction performance, improving upon the best baseline by 21.30 dB on the Kodak benchmark. Future work will explore extending JA-SIREN to data compression, neural codec for videos, and novel view synthesis.

\vspace{-10pt}

\newcommand{\cem}[1]{\textcolor{blue}{cem: #1}}

\bibliographystyle{IEEEbib}
\bibliography{strings,refs}

@String(CVPR  = {IEEE Conf. Comput. Vis. Pattern Recog.})

@String(NeurIPS = {Adv. Neural Inform. Process. Syst.})

@String(JMLR  = {J. Mach. Learn. Res.})

@String(CVPR  = {CVPR})

@String(NeurIPS = {NeurIPS})

@String(JMLR  = {JMLR})

@inproceedings{sitzmann2019siren,
author = {Sitzmann, Vincent
          and Martel, Julien N.P.
          and Bergman, Alexander W.
          and Lindell, David B.
          and Wetzstein, Gordon},
title = {Implicit Neural Representations
          with Periodic Activation Functions},
booktitle = {Proc. NeurIPS},
year={2020}
}

@article{ahmed2006discrete,
  title={Discrete cosine transform},
  author={Ahmed, Nasir and Natarajan, T\_ and Rao, Kamisetty R},
  journal={IEEE transactions on Computers},
  volume={100},
  number={1},
  pages={90--93},
  year={2006},
  publisher={IEEE}
}

@article{essakine2024we,
  title={Where do we stand with implicit neural representations? a technical and performance survey},
  author={Essakine et al.},
  journal={arXiv preprint arXiv:2411.03688},
  year={2024}
}

@article{tancik2020fourier,
  title={Fourier features let networks learn high frequency functions in low dimensional domains},
    author={Tancik, Matthew and others},
    journal={Advances in neural information processing systems},
  volume={33},
  year={2020}
}

@article{mildenhall2021nerf,
  title={Nerf: Representing scenes as neural radiance fields for view synthesis},
  author={Mildenhall, Ben and Srinivasan, Pratul P and Tancik, Matthew and Barron, Jonathan T and Ramamoorthi, Ravi and Ng, Ren},
  journal={Communications of the ACM},
  volume={65},
  number={1},
  year={2021},
  publisher={ACM New York, NY, USA}
}

@inproceedings{saragadam2023wire,
  title={WIRE: Wavelet Implicit Neural Representations},
    author={Saragadam, Vishwanath and others},  
    booktitle={Conf. Computer Vision and Pattern Recognition},
  year={2023}
}

@inproceedings{ramasinghe2022beyond,
  title={Beyond periodicity: Towards a unifying framework for activations in coordinate-mlps},
  author={Ramasinghe, Sameera and Lucey, Simon},
  booktitle={European Conference on Computer Vision},
  pages={142--158},
  year={2022},
  organization={Springer}
}

@inproceedings{liu2024finer,
  title={Finer: Flexible spectral-bias tuning in implicit neural representation by variable-periodic activation functions},
    author={Liu, Zhen and others},
    booktitle={Proceedings of the IEEE/CVF Conference on Computer Vision and Pattern Recognition},
  year={2024}
}

@inproceedings{spokenwikipedia,
  author = {Arne K{\"o}hn and Florian Stegen and Timo Baumann},
  title = {Mining the Spoken {Wikipedia} for Speech Data and Beyond},
  booktitle = {Proc. LREC},
  year = {2016},
  address = {Portoro{\v{z}}, Slovenia},
  publisher = {ELRA}
}

@misc{kodak,
  author = {Mehta, Sheryl},
  title = {{Kodak Lossless True Color Image Suite}},
  year = {2020},
  publisher = {Kaggle},
  url = {https://www.kaggle.com/datasets/sherylmehta/kodak-dataset},
  urldate = {2025-08-28}
}

@misc{usc-sipi-database,
  title        = {USC-SIPI Image Database},
  author       = {{USC Signal and Image Processing Institute}},
  howpublished = {\url{https://sipi.usc.edu/database/}},
  year         = {1973}
}

@inproceedings{shi2024improved,
  title={Improved implicit neural representation with fourier reparameterized training},
  author={Shi, Kexuan and Zhou, Xingyu and Gu, Shuhang},
  booktitle={Proceedings of the IEEE/CVF Conference on Computer Vision and Pattern Recognition},
  pages={25985--25994},
  year={2024}
}

@inproceedings{jayasundara2025mire,
  title={MIRE: Matched Implicit Neural Representations},
  author={Jayasundara, Dhananjaya and Zhao, Heng and Labate, Demetrio and Patel, Vishal M},
  booktitle={Proceedings of the Computer Vision and Pattern Recognition Conference},
  pages={8279--8288},
  year={2025}
}

@misc{philips_circle_wiki,
  title        = {Philips Circle Pattern},
  author       = {{Wikipedia contributors}},
  year         = {2025}
}

@article{bottou2018optimization,
  title={Optimization methods for large-scale machine learning},
  author={Bottou, L{\'e}on and Curtis, Frank E and Nocedal, Jorge},
  journal={SIAM review},
  volume={60},
  number={2},
  year={2018},
  publisher={SIAM}
}

@inproceedings{shah2024spder,
  title={Spder: Semiperiodic Damping-Enabled Object Representation},
  author={Shah, Kathan and Sitawarin, Chawin},
  booktitle={International Conference on Learning Representations},
  year={2024}
}

@article{kania2024fresh,
  title={Fresh: Frequency shifting for accelerated neural representation learning},
author={Kania, Adam and others},
  journal={arXiv preprint arXiv:2410.05050},
  year={2024}
}

@inproceedings{novello2025tuning,
  title={Tuning the frequencies: Robust training for sinusoidal neural networks},
  author={Novello, Tiago and Aldana, Diana and Araujo, Andre and Velho, Luiz},
  booktitle={Proceedings of the Computer Vision and Pattern Recognition Conference},
  year={2025}
}

@inproceedings{sitzmann2019scene,
  title     = {Scene Representation Networks: Continuous 3D-Structure-Aware Neural Scene Representations},
  author    = {Sitzmann, Vincent and others},
  booktitle = {Advances in Neural Information Processing Systems},
  volume    = {32},
  year      = {2019}
}

@article{martel2021acorn,
  title={Acorn: Adaptive coordinate networks for neural scene representation},
  author={Martel, Julien NP and Lindell, David B and Lin, Connor Z and Chan, Eric R and Monteiro, Marco and Wetzstein, Gordon},
  journal={arXiv preprint arXiv:2105.02788},
  year={2021}
}

@inproceedings{strumpler2022implicit,
  title={Implicit neural representations for image compression},
  author={Str{\"u}mpler, Yannick and Postels, Janis and Yang, Ren and Gool, Luc Van and Tombari, Federico},
  booktitle={European conference on computer vision},
  pages={74--91},
  year={2022},
  organization={Springer}
}

@article{paszke2019pytorch,
  title={Pytorch: An imperative style, high-performance deep learning library},
  author={Paszke et al.},
  journal={Advances in neural information processing systems},
  volume={32},
  year={2019}
}

@inproceedings{atzmon2020sal,
  title={Sal: Sign agnostic learning of shapes from raw data},
  author={Atzmon, Matan and Lipman, Yaron},
  booktitle={Proceedings of the IEEE/CVF conference on computer vision and pattern recognition},
  pages={2565--2574},
  year={2020}
}

@inproceedings{hore2010image,
  title={Image quality metrics: PSNR vs. SSIM},
  author={Hore, Alain and Ziou, Djemel},
  booktitle={2010 20th international conference on pattern recognition},
  year={2010},
  organization={IEEE}
}

@InProceedings{Ben-Shabat_2022_CVPR,
    author    = {Ben-Shabat, Yizhak and Koneputugodage, Chamin Hewa and Gould, Stephen},
    title     = {DiGS: Divergence Guided Shape Implicit Neural Representation for Unoriented Point Clouds},
    booktitle = {Proceedings of the IEEE/CVF Conference on Computer Vision and Pattern Recognition (CVPR)},
    month     = {June},
    year      = {2022},
    pages     = {19323-19332}
}

@inproceedings{glorot2010understanding,
  title={Understanding the difficulty of training deep feedforward neural networks},
  author={Glorot, Xavier and Bengio, Yoshua},
  booktitle={Proceedings of the thirteenth international conference on artificial intelligence and statistics},
  pages={249--256},
  year={2010},
  organization={JMLR Workshop and Conference Proceedings}
}

@inproceedings{he2015delving,
  title={Delving deep into rectifiers: Surpassing human-level performance on imagenet classification},
  author={He, Kaiming and Zhang, Xiangyu and Ren, Shaoqing and Sun, Jian},
  booktitle={Proceedings of the IEEE international conference on computer vision},
  pages={1026--1034},
  year={2015}
}

@article{alsakabi2025fm,
  title={FM-SIREN \& FM-FINER: Implicit Neural Representation Using Nyquist-based Orthogonality},
  author={Alsakabi, Mohammed and Mobeirek, Wael and Dolan, John M and Tonguz, Ozan K},
  journal={arXiv preprint arXiv:2509.23438},
  year={2025}
}

@article{dattoli1990theory,
  title={Theory of generalized Bessel functions},
  author={Dattoli, G and Giannessi, L and Mezi, L and Torre, A},
  journal={Il Nuovo Cimento B (1971-1996)},
  volume={105},
  number={3},
  pages={327--348},
  year={1990},
  publisher={Springer}
}

@misc{jacobianger,
  author       = {Weisstein, Eric W.},
  title        = {Jacobi-Anger Expansion},
  howpublished = {MathWorld--A Wolfram Resource},
  url          = {https://mathworld.wolfram.com/Jacobi-AngerExpansion.html}
}

@inproceedings{koneputugodage2025vi,
  title={VI\^{} 3NR: Variance Informed Initialization for Implicit Neural Representations},
  author={Koneputugodage, Chamin Hewa and Ben-Shabat, Yizhak and Ramasinghe, Sameera and Gould, Stephen},
  booktitle={Proceedings of the Computer Vision and Pattern Recognition Conference},
  pages={13477--13486},
  year={2025}
}

@article{xu2022signal,
  title={Signal processing for implicit neural representations},
  author={Xu, Dejia and Wang, Peihao and Jiang, Yifan and Fan, Zhiwen and Wang, Zhangyang},
  journal={Advances in Neural Information Processing Systems},
  volume={35},
  pages={13404--13418},
  year={2022}
}

@article{patel2025normal,
  title={Normal-guided Detail-Preserving Neural Implicit Function for High-Fidelity 3D Surface Reconstruction},
  author={Patel, Aarya and Laga, Hamid and Sharma, Ojaswa},
  journal={Proceedings of the ACM on computer graphics and interactive techniques},
  volume={8},
  number={1},
  pages={1--24},
  year={2025},
  publisher={ACM New York, NY}
}

\end{document}